\documentclass[10pt,twocolumn,letterpaper]{article}

\usepackage[pagenumbers]{cvpr} 
\usepackage[utf8]{inputenc} 
\usepackage[T1]{fontenc}    
\usepackage{url}            
\usepackage{booktabs}       
\usepackage{amsfonts}       
\usepackage{nicefrac}       
\usepackage{microtype}      
\usepackage{xcolor}         
\usepackage{graphicx}
\usepackage{multirow}
\usepackage{makecell} 
\usepackage{subcaption}

\definecolor{cvprblue}{rgb}{0.21,0.49,0.74}
\usepackage[pagebackref,breaklinks,colorlinks,allcolors=cvprblue]{hyperref}

\def\paperID{*****} 
\def\confName{CVPR}
\def\confYear{2025}

\title{Dynamic Graph Induced Contour-aware Heat Conduction Network for Event-based Object Detection}

\author{Xiao Wang$^{1}$, Yu Jin$^{1}$, Lan Chen$^{2}$\thanks{Corresponding Author: Lan Chen \& Bo Jiang}, Bo Jiang$^{1}$*, Lin Zhu$^{3}$, Yonghong Tian$^{4,5,6}$, Jin Tang$^{1}$, Bin Luo$^{1}$ \\ 
${^1}${School of Computer Science and Technology, Anhui University, Hefei, China} \\
${^2}${School of Electronic and Information Engineering, Anhui University, Hefei, China} \\
${^3}${Beijing Institute of Technology, Beijing, China} \\
${^4}${Peng Cheng Laboratory, Shenzhen, China} \\ 
${^5}${School of Computer Science, Peking University, China} \\ 
${^6}${School of Electronic and Computer Engineering, Shenzhen Graduate School, Peking University, China} \\ 
\small{\textit{jy0x4f@163.com}, \textit{\{xiaowang, chenlan, jiangbo, tangjin, luobin\}@ahu.edu.cn}, \textit{\{linzhu, yhtian\}@pku.edu.cn}}  
}

\begin{document}
\maketitle

\begin{abstract}
Event-based Vision Sensors (EVS) have demonstrated significant advantages over traditional RGB frame-based cameras in low-light conditions, high-speed motion capture, and low latency. Consequently, object detection based on EVS has attracted increasing attention from researchers. Current event stream object detection algorithms are typically built upon Convolutional Neural Networks (CNNs) or Transformers, which either capture limited local features using convolutional filters or incur high computational costs due to the utilization of self-attention. Recently proposed vision heat conduction backbone networks have shown a good balance between efficiency and accuracy; however, these models are not specifically designed for event stream data. They exhibit weak capability in modeling object contour information and fail to exploit the benefits of multi-scale features. To address these issues, this paper proposes a novel dynamic graph induced contour-aware heat conduction network for event stream based object detection, termed CvHeat-DET. The proposed model effectively leverages the clear contour information inherent in event streams to predict the thermal diffusivity coefficients within the heat conduction model, and integrates hierarchical structural graph features to enhance feature learning across multiple scales. Extensive experiments on three benchmark datasets for event stream-based object detection fully validated the effectiveness of the proposed model. 
The source code of this paper will be released on \url{https://github.com/Event-AHU/OpenEvDET} 
\end{abstract}

\section{Introduction} 
Object detection targets locate the object we are interested in and provide the bounding box and category label of each object. It is a basic research problem in the computer vision community and helps other high-level tasks, such as image captioning~\cite{hossain2025cm, li2025m3ixup, cai2025toward}, object tracking~\cite{wang2025omnitracker,shi2025mambatrack, hu2025exploiting, chen2025sutrack}, scene graph generation~\cite{jiang2025enhancing, tang2025remote, wang2025indvissgg, liu2025relation}. Object detectors can be widely used in many practical scenarios, including intelligent video surveillance, robotics, etc. However, the mainstream frame-based object detection is easily influenced by low illumination and fast motion, thus, the object detection in challenging scenarios is still unsatisfying.

\begin{figure*} 
\centering
\includegraphics[width=0.85\linewidth]{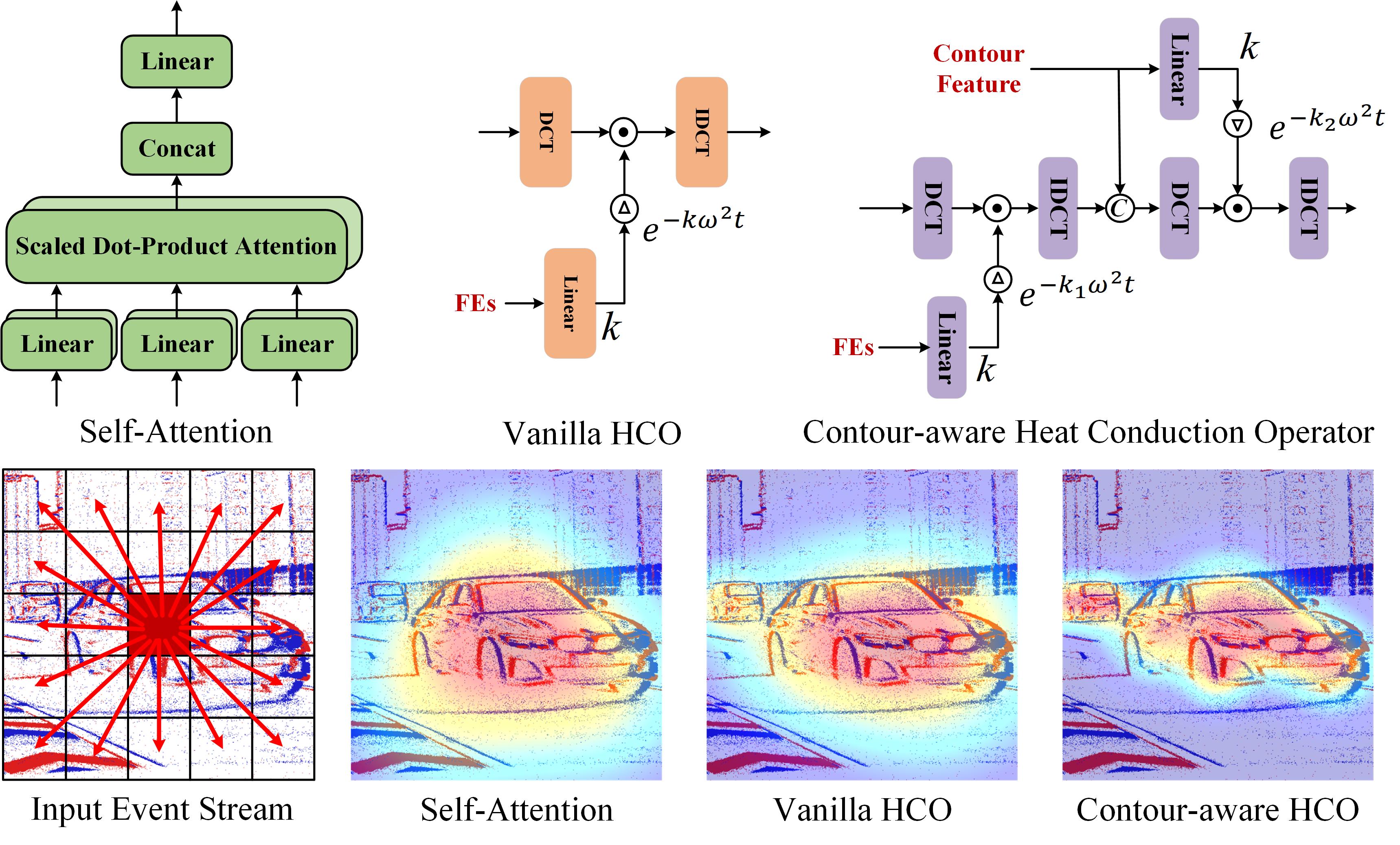}
\caption{ 
(a). Architectures of Self-Attention, Vanilla Heat Conduction Operator (HCO), and our proposed Contour-aware Heat Conduction Operator (HCO). 
(b). Information Conduction Mechanisms of Self-Attention, Vanilla Heat Conduction (HCO), and our proposed Contour-aware HCO.} 
\label{fig_overview}
\end{figure*}

In recent years, the Event-based Vision Sensors (EVS, also called event camera) have drawn more and more attention and have been widely used in object detection~\cite{gehrig2023rvt, su2023emsyolo, luo2024spikeyolo, guo2024stat, peng2024sast, wang2024mvheatdet}, visual object tracking~\cite{wang2024eventtrack1, tang2022eventtrack2, wang2024eventtrack3, wang2024mambaevt}, pattern recognition~\cite{chen2024velora, wang2023sstformer, yuan2023learning}, and video captioning~\cite{wang2024eventcaption1, wang2025signcaption}. Because it performs better on high dynamic range, high temporal resolution, low energy consumption, and low latency, as noted in~\cite{gallego2020eventcamera}. The event camera emits a spike signal if and only if the brightness change exceeds a specific threshold, and events from different pixels are generated asynchronously, unlike RGB cameras which capture scene brightness information through global exposure. The event stream is recorded line by line in the form of $(x, y, t, p)$ tuples, where $(x, y)$ represents the spatial location, and $t$ and $p$ denote the timestamp and the polarity of the event, respectively. 
For the event-based detection we studied in this paper, Gehrig et al. propose RVT~\cite{gehrig2023rvt}, a novel architecture that synergistically integrates attention mechanisms with LSTM to achieve an optimal balance between computational accuracy and processing speed. 
Wang et al. propose vHeat~\cite{wang2024vheat}, represents the first successful integration of thermal conduction from physics into computer vision, achieving state-of-the-art performance in image classification, object detection, and semantic segmentation tasks.
Wang et al. propose MvHeat-DET~\cite{wang2024mvheatdet}, which introduces MoE framework to adaptively process data with diverse characteristics, and build a high-resolution event-based object detection benchmark dataset.

Despite the significant progress of the event-based object detection, however, these models may still be limited by the following issues: 
1). Current event stream based detection algorithms generally adopt CNN~\cite{wang2023dual, cho2025ev, lu2024flexevent, cao2024embracing} or ViT models~\cite{gehrig2023rvt, torbunov2024evrtdetr, peng2024sast, peng2023get, guo2024stat} as the backbone, however, these backbone networks either capture only local information or involve high computational complexity, failing to fully exploit the dense temporal resolution of event streams. 
2). Some works attempt to achieve a better trade-off between computational cost and performance from the perspective of heat conduction models~\cite{wang2024vheat, wang2024mvheatdet}, however, these algorithms are directly adapted to event stream data without fully exploiting the unique characteristics of event streams, namely, their clear edge and contour information.
Moreover, existing methods employ only a single representation of the event stream for feature learning, thus failing to take full advantage of multi-scale/-view feature representations. 
Accordingly, a natural question arises: \textit{"How can we effectively incorporate the intrinsic contour cues of event streams into the design of a heat conduction-based backbone, while simultaneously exploiting hierarchical multi-scale features to enhance event stream representation?"}

To address the aforementioned challenges, this paper proposes a novel dynamic graph induced contour-aware heat conduction network for event-based object detection, referred to as CvHeat-DET. The key insight of our work lies in structurally parsing event streams using a dynamic graph neural network, thereby fully exploiting the structural characteristics of contours in the event data. This significantly enhances the heat conduction model's ability to represent and model event streams. During this process, structural features are effectively integrated into the heat conduction framework, improving the model's capacity for multi-scale/-view feature learning. Specifically, we first adopt a stem network to encode the input event streams, then, feed them into the contour HCO (Heat Conduction Operation) layers for balanced accuracy and computational cost. Meanwhile, dynamic graph-induced contour-aware feature learning is employed to assist in predicting the thermal diffusivity coefficients in the heat conduction model, thereby enabling a better adaptation of HCO to the event domain. Additionally, this hierarchical feature representation is directly integrated into the heat conduction layer to achieve multi-scale and multi-modal feature encoding. Finally, we adopt an IoU-based query selection module for feature selection and detect the target object using a detection head. An overview of our proposed CvHeat-DET framework can be found in Fig.~\ref{fig_overview}.

To sum up, the contributions of this work can be summarized as the following three aspects: 

1). We propose a novel contour-guided heat conduction framework for accurate event-based object detection, termed CvHeat-DET.  

2). We propose a novel dynamic structured graph to learn the edge cues from the event streams and guide the object detector from hierarchical heat conduction and multi-modal fusion. 

3). Extensive experiments conducted on multiple event stream-based object detection benchmark datasets fully validated the effectiveness of our proposed model.

\section{Related Works} 
\noindent \textbf{Event-based Object Detection.~} 
Event cameras, due to their high temporal resolution, are particularly well-suited for detecting moving objects. 
EvRT-DETR~\cite{torbunov2024evrtdetr}, STAT~\cite{guo2024stat}, SAST~\cite{peng2024sast}, RVT~\cite{gehrig2023rvt} and GET~\cite{peng2023get} utilize Transformer-based architectures as the backbone for feature extraction. Specifically, EvRT-DETR adapts RT-DETR for real-time object detection on event data; STAT employs an adaptive selection mechanism to extract high-quality event representations; and SAST, GET, and RVT integrate LSTM modules to further exploit the high temporal resolution of event data.
Due to the unique imaging mechanism of event cameras, many current studies tend to employ SNNs for event-based object detection. SpikingYOLO~\cite{kim2020spikingyolo}, SpikeYOLO~\cite{luo2024spikeyolo}, EMS-YOLO~\cite{su2023emsyolo} and HDI-Former~\cite{li2024hdiformer} utilize SNNs as backbones to process asynchronous event data. In particular, SpikeYOLO introduces I-LIF (Integer Leaky Integrate-and-Fire) neurons to mitigate quantization errors in SNNs, significantly narrowing the performance gap between SNNs and conventional CNNs in object detection tasks. 
Recently, Wang et al. proposed a novel foundational model, vHeat~\cite{wang2024vheat}, to alleviate the high computational complexity of attention mechanisms. Building upon this, MvHeat-DET~\cite{wang2024mvheatdet} applies the method to event data and incorporates a MoE mechanism, achieving impressive performance in object detection tasks. 
Different from these works, in this paper, we propose a new contour-guided vision heat conduction model for event-based object detection.  It simultaneously achieves contour-guided thermal diffusivity prediction in the vision HCO model and multi-scale feature fusion, significantly improving the performance of event stream object detection.

\noindent \textbf{Graph Neural Networks.~}   
Graph Neural Networks (GNNs) are a class of deep learning models specifically designed to handle graph-structured data. They can model arbitrary relational structures between nodes and can be flexibly integrated with models such as CNNs and Transformers to enhance performance in multi-modal or structured tasks.
Some existing works~\cite{lanchantin2020gnnclassification1, wu2020gnnclassification2, ye2020gnnclassification3, nguyen2021gnnclassification4, zhao2021gnnclassification5} combine graph neural networks with Transformers or CNNs to model semantic label dependencies for classification tasks.
The works~\cite{zhang2019gnnseg1, hu2020gnnseg2, zhang2019gnnseg3, zhang2021gnnseg4, wu2020gnnseg5} employ GNNs to perform image segmentation by effectively capturing the structured relationships between pixels or regions, achieving a favorable balance between accuracy and speed. 
For the object detection task, FGRR~\cite{chen2022daod} constructs a graph to model intra-domain and inter-domain foreground object relations. GraphFPN~\cite{zhao2021graphfpn} combines adaptive topological structures with attention mechanisms in GNN to enhance multi-scale feature learning capability. 
Due to the natural alignment between the ability of Graph Neural Networks (GNNs) to process unstructured data and the sparsity and asynchrony of event data, an increasing number of studies are integrating GNNs with event data processing. DAGr~\cite{gehrig2024dagr} uses asynchronous GNN to construct spatio-temporal graphs from events, achieving a favorable bandwidth–latency trade-off. AEGNN~\cite{schaefer2022aegnn} models events as evolving spatio-temporal graphs for efficient classification and detection. 
Inspired by these works, in this paper, we propose to model the contour cues of event streams using a structured dynamic graph, which further enhances the object detection task.

\section{Methodology} 

\subsection{Overview} 
Fig.~\ref{fig_overview} illustrates the overall architecture of our proposed CvHeat-DET framework. The input to the network is event frames, and the entire framework can be viewed as consisting of two branches: the \textit{Dynamic Graph Network} and \textit{Contour-aware Heat Conduction Network}. First, the event frame is fed into the graph branch to construct graph structures. Through operations such as splitting and filtering, graphs with varying numbers of nodes are generated. These graphs are then processed using Graph Convolutional Networks (GCNs) to extract features at multiple scales, which are subsequently fed into the backbone network. Then, event frame is first divided into patches and extracted low-level features through a lightweight stem network. These features are then fed into a four-stage heat conduction module. This module fuses graph features with event features, and leverages the graph features to constrain the network's attention within the interior regions of the target objects, thereby improving detection accuracy.

\begin{figure*} 
\centering
\includegraphics[width=\linewidth]{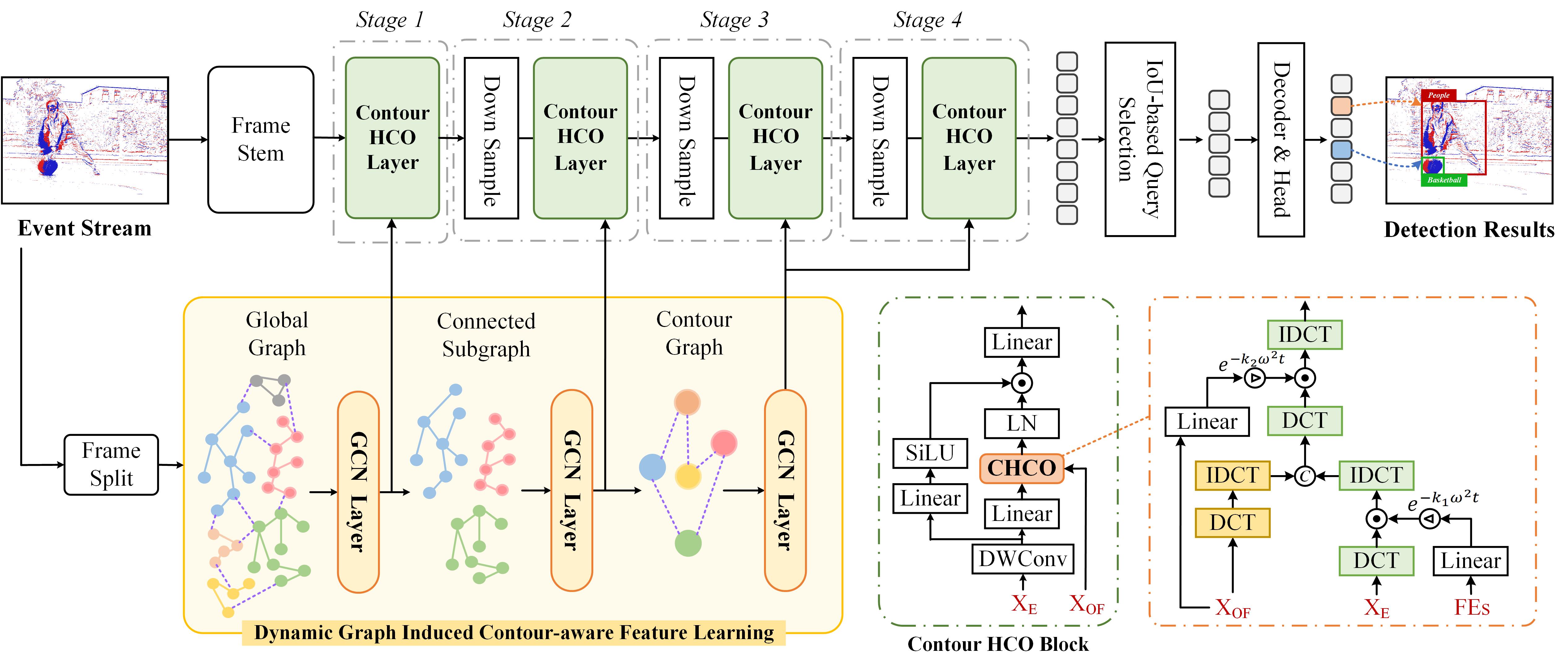}
\caption{An overview of our proposed event-based object detection framework, termed CvHeat-DET.}
\label{fig_overview}
\end{figure*}

\subsection{Input Encoding} 
Event-based cameras~\cite{gallego2020eventcamera} are bio-inspired sensors that capture event streams in an asynchronous and sparse manner. The asynchronous data captured by event cameras can be represented as $\mathcal{E} = (x_i, y_i, t_i, p_i), i \in N$, where $x_i$ and $y_i$ denote the spatial coordinates, $t_i$ represents the timestamp of the event, and $p_i \in \{-1, 1\}$ indicates the polarity, where 1 represents an increase in pixel brightness and -1 represents a decrease. $N \in \{1, 2, 3, \cdots, \mathcal{N} \}$ denotes the index of the event.

The traditional event-to-representation conversion process can be divided into two steps~\cite{cao2024spikeslicer}:  
1) slicing the raw event stream into sub-event streams based on specific rules (e.g., fixed time intervals or a fixed number of events);  
2) converting each sub-event stream into a specific event representation.
In this work, we choose to slice the event stream at fixed time intervals and convert each sub-event stream into an event frame.

\subsection{CvHeat-DET: Contour-aware Visual Heat Conduction for Event-based Detection} 
Traditional object detection algorithms typically adopt CNNs or Transformers as backbone networks. However, these approaches often suffer from limited accuracy or high computational cost. Recently, heat conduction-based models have gained significant attention due to their $\mathcal{O}(N^{1.5})$ complexity. These models first apply a frequency domain transformation function $\mathcal{F}(\cdot)$ to the input image, then predict a thermal diffusivity coefficient $k$, and perform computations based on the transformed representation. Finally, the result is mapped back to the original domain using an inverse transformation $\mathcal{F}^{-1}(\cdot)$. Heat conduction models have demonstrated promising performance in RGB image detection tasks, offering a compelling trade-off between accuracy and efficiency.

Event cameras possess high temporal resolution, high dynamic range, and low latency, enabling them to asynchronously record pixel-level brightness changes at microsecond precision. This unique sensing mechanism makes them particularly effective at capturing contour information in dynamic scenes. We propose that incorporating object contour information into the heat diffusion mechanism can effectively guide the heat flow along object boundaries. This not only helps preserve critical structural details but also suppresses background interference. To achieve this goal, we employ graph structures to model object contours, as they can flexibly represent irregular connectivity and adapt to complex boundary shapes and structural features.

Moreover, in object detection tasks, single-scale features are often insufficient to effectively capture targets of varying sizes. Integrating features from multiple scales enhances the model’s ability to perceive objects at different scales, leading to more robust performance in complex scenes and effectively reducing missed and false detections. To this end, we preserve multi-scale graph representations with diverse structural characteristics during the graph construction process. These representations are then processed by a GCN to extract multi-scale features, which further help the model precisely control the diffusion of information along object contours.

\noindent $\bullet$ \textbf{Dynamic Graph Induced Contour-aware Feature Learning}
In object detection tasks, graph structures are effective in modeling non-local relationships between objects, which is especially beneficial in complex scenes for improving detection accuracy. In this work, we propose to model object contours using graph structures. By incorporating multi-scale graph features, our method achieves significant improvements in detection performance.

\noindent \textbf{Global Graph Construction.}
Given the input event frame $\mathcal{I} \in \mathbb{R}^{C \times H \times W}$, firstly, we divide it into non-overlapping patches of fixed size. The patches $p_i \in \mathbb{R}^{C \times H_p \times W_p}$, where $H_p$ and $W_p$ denote the height and width of each patch, respectively, $i \in \{1, 2, 3,\cdots,\frac{H}{H_p}\times\frac{W}{W_p} \}$. Each patch is treated as a graph node \( v_i \), with the patch's center coordinates used as the node position and the entire image data within the patch used as the node feature for graph construction. Thus, the nodes of global graph can be denoted as $V = \{v_1, v_2, v_3, \cdots,v_n\}$, where $n = \frac{H}{H_p}\times\frac{W}{W_p}$. To establish the graph edge $E$, we compute the Euclidean distance between nodes. If the distance is less than a predefined distance threshold $R_d$, an edge is established between the nodes; otherwise, no edge is formed. Thus, the global graph $G_g = (V, E)$ is established, effectively preserving the rich spatial information contained within the event frames, aiding the shallow heat conduction stage in better capturing the global information.

\noindent \textbf{Connected Subgraphs Construction.}
Event cameras only produce data when pixel brightness changes and when using an event camera to capture moving objects, the contour of the objects can be clearly observed, while the details inside the contours appear blurry and sparse~\cite{wu2024egsst}. Consequently, the generated data exhibits a characteristic pattern: events are dense in the areas corresponding to object contours, whereas they are sparse in the background and inside the contours. When such characteristic event data is used to construct a graph, the the regions with dense events contain more nodes and edges, typically representing the object contours. 

In detection tasks, attention needs to be focused solely on the object contours, while the background and internal details of the objects can be disregarded. This approach facilitates more efficient target recognition and analysis while reducing unnecessary computational load. We utilize the Louvain algorithm~\cite{blondel2008louvain} to partition the global graph into $K$ connected subgraphs, $G_k = (V_k, E_k)$, which satisfy
\begin{equation}
\label{eq_subgraph}
    D = \{ G_k: G_k \in G_g, k = 1,2,3, \cdots, K\}.
\end{equation} 
If the number of nodes in a subgraph exceeds a predefined node threshold $R_n$, it is considered to model the contour information of an object. Conversely, subgraphs with fewer nodes are deemed to represent background or detailed information. By filtering out subgraphs with fewer than $R_n$ nodes, finally, we get the connected subgraphs $D_f \subseteq D$, and realize downsampling.

\noindent \textbf{Contour Graph Construction.} 
Since the spatial features of nodes within each subgraph $G_k$ are similar, retaining all node features of the subgraph is redundant~\cite{wu2024egsst}. To reduce the computational burden, we aggregate each subgraph $G_k$ into a single node $v_k$ as follows:
\begin{equation}
\label{eq_aggregate}
    \mathcal{V} = \{ v_k = \pi(V_k), k=1, 2, 3, \cdots, K\},
\end{equation}
where $\pi$ is a function for aggregation through spatial coordinates and node features, we use the mean function to better preserve the overall information and trends of the original data. We then use the K-nearest neighbors algorithm~\cite{cover1967knn} to connect these aggregate nodes, creating a new edges $\mathcal{E}$. Thus, we create a new graph $\mathcal{G} = \{ \mathcal{V}, \mathcal{E}\}$ with fewer nodes and edges, named \textit{contour graph}, which significantly decreases computational complexity and processing delay.

We employ different GCNs to extract features from graphs at various scales ($G_g, D_l, \mathcal{G}$). These features are then fed into the heat conduction network, enhancing the network to focus on the characteristics of object parts more effectively.

\noindent $\bullet$ \textbf{Contour-aware Visual Heat Conduction Network}  
Inspired by vHeat~\cite{wang2024vheat}, we employ the Discrete Cosine Transform (DCT) to implement the function $\mathcal{F}$ and inverse Discrete Cosine Transform (IDCT) to implement $\mathcal{F}^{-1}$.

The Heat Conduction Network consists of four stages, each comprising a downsampling layer followed by a Contour Heat Conduction Operator (CHCO) layer, and each CHCO layer is stacked with several CHCO blocks. The input of the CHCO block is event features $X_E$ from the last stage and contour graph feature $X_{OF}$ from the Dynamic Graph Network. For the CHCO block, the input feature $X_E$ is first processed by DWConv and a linear transformation, then combined with $X_{OF}$ and fed into the CHCO module for feature fusion and heat conduction modeling. The output is subsequently added to the transformed $X_E$ via a residual connection, and finally passed through a linear layer to produce the fused features. The CHCO block is similar to the ViT block while replacing self-attention operators with CHCOs.

For the CHCO, the event features $X_E$ undergoing a standard heat conduction process. The thermal diffusivity coefficient $k_1$ is adaptively learned from the frequency embeddings (FEs) through a linear layer. Then, $X_E$ is transformed into the frequency domain via the DCT, multiplied by the coefficient matrix $e^{-k_1w^2t}$, and finally converted back to the spatiotemporal domain using IDCT, yielding the enhanced heat features $X_f$. Then, we concatenate the contour features $X_{OF}$ with the enhanced features $X_f$ to fuse the information, which enhances the model’s ability to perceive object contours and dynamic regions. We apply a DCT to $X_{OF}$, followed immediately by an IDCT to recover it. This process facilitates frequency-domain reconstruction and alignment, enhancing the consistency and representational capacity for subsequent feature fusion. Finally, we apply a second heat conduction process to the fusion feature, where the thermal diffusivity coefficient $k_2$ is predicted from the original contour features. This allows the model to adaptively modulate the diffusion strength based on structural information, enabling more precise and contour-aware feature propagation, achieving the effect of confining heat within object boundaries.

\begin{table*}[!h]
\centering
\caption{Experimental results on EvDET200K benchmark dataset. 
    vHeat$^{\dagger}$ means using vHeat as the encoder and Transformer as the decoder.}
    \label{tab:EvDET200Kresults}
    \resizebox{\textwidth}{!}{ 
    \begin{tabular}{c|l|l|l|c|c|c|c|c|c|c}
    \toprule 
    \textbf{Index} &\textbf{Algorithm} &\textbf{Publish}  &\textbf{Backbone}  &\textbf{mAP@50:95} &\textbf{mAP@50} &\textbf{mAP@75} &\textbf{Params} &\textbf{FLOPs} &\textbf{FPS} &\textbf{Code} \\
    \hline  
    01   &Faster R-CNN~\cite{girshick2015fasterrcnn}  &TPAMI 2016  &ResNet50  &46.0 &73.3 &48.6   &40.9M &71.2G  &23
    &\href{https://github.com/rbgirshick/py-faster-rcnn}{URL} \\      
    \hline     
    02   &S5-ViT~\cite{zubic2024s5vit}  &CVPR 2024  &Former+SSM  &42.9 &76.3 &44.1  &18.2M &5.6G  &84 
    &\href{https://github.com/uzh-rpg/ssms_event_cameras}{URL} \\ 
    \hline  
    03   &SpikeYOLO~\cite{luo2024spikeyolo}  &ECCV 2024  &SNN  &41.2  &74.8 &39.8  &68.8M &78.1G  &77  &\href{https://github.com/BICLab/SpikeYOLO}{URL} \\
    \hline      
    \multirow{4}{*}{04} & YOLOv10-N~\cite{wang2024yolov10} & \multirow{4}{*}{NeurIPS 2024} &\multirow{4}{*}{CNN}  &42.7 &75.1 &42.1 &2.3M &8.2G &116 &\multirow{4}{*}{\href{https://github.com/THU-MIG/yolov10}{URL} }\\
    &YOLOv10-S~\cite{wang2024yolov10} & & &43.5 &76.2 &43.0 &7.3M  &21.6G &83 &\\
    &YOLOv10-M~\cite{wang2024yolov10} & & &44.0 &77.5 &42.8 &15.4M &59.1G &32  &\\
    &YOLOv10-B~\cite{wang2024yolov10} & & &44.1 &77.9 &43.1 &19.1M  &92.0G &30 &\\
    \hline  
    05   &RVT~\cite{gehrig2023rvt}  &CVPR 2023  &Transformer  &40.7 &73.1 &42.3 & 9.9M &8.4G &88 
    &\href{https://github.com/uzh-rpg/RVT}{URL} \\
    \hline  
    06   &EMS-YOLO~\cite{su2023emsyolo}  &ICCV 2023  &SNN  &32.1  &66.6 &27.4 &14.40M &3.3M  &119 
    &\href{https://github.com/BICLab/EMS-YOLO}{URL} \\
    \hline  
    07   &YOLOv6~\cite{li2022yolov6}  &arXiv 2022  &RepVGG  &41.3  &75.7 &38.9 &17.2M &44.2M  &70
    &\href{https://github.com/meituan/YOLOv6}{URL} \\
    \hline  
    08   &Swin-T~\cite{liu2021swintransformer}  &ICCV 2021  &Transformer  &49.0 &78.4 &52.7  &160M &1043G  &26 &\href{https://github.com/microsoft/Swin-Transformer}{URL} \\
    \hline  
    09   &DetectoRS~\cite{qiao2021detectors}  &CVPR 2021  &ResNet50  &49.1 &78.8 &53.5 &123.2M &117.2G  &32
    &\href{https://github.com/joe-siyuan-qiao/DetectoRS}{URL} \\
    \hline  
    10   &RED~\cite{perot2020red1mpx} &NeurIPS 2020  &ResNet50  &35.4 &68.3 &35.2 &24.1M &46.3G  &34
    &\href{https://github.com/prophesee-ai/prophesee-automotive-dataset-toolbox}{URL} \\
    \hline  
    11   &DETR~\cite{carion2020detr}  &ECCV 2020  &Transformer  &40.9 &74.5 &39.5 &41M &86G  &29 &\href{https://github.com/facebookresearch/detr}{URL} \\
    \hline  
    12   &Mask R-CNN~\cite{he2017maskrcnn}  &ICCV 2017  &ResNet50  &48.8 &77.6 &52.0 &43.8M &142.7G  &28
    &\href{https://github.com/yhenon/pytorch-retinanet}{URL} \\
    \hline  
    13   &RetinaNet~\cite{lin2017retinanetfocal}  &ICCV 2017  &ResNet50  &48.6 &77.0 &50.8 &36.2M &81.4G  &76 &\href{https://github.com/yhenon/pytorch-retinanet}{URL} \\
    \hline
    14   &vHeat$^{\dagger}$~\cite{wang2024vheat}  &CVPR 2025  &HCO  &50.3 &72.2 &54.2 &56.3M &74.5G &50 
    &\href{https://github.com/uzh-rpg/ssms_event_cameras}{URL} \\    
    \hline      
    15   &MvHeat-DET~\cite{wang2024mvheatdet}  &CVPR 2025  &MHCO  &52.9      &80.4       &55.9      &47.5M       &56.4G       &58    &\href{https://github.com/Event-AHU/OpenEvDET}{URL}  \\ 
    \hline   
    16   &CvHeat-DET (Ours)  &-  &CHCO  &\textbf{53.6}      &\textbf{80.9}       &\textbf{56.9}       &76.0M      &98.8G     &11    &-   \\ 
    \bottomrule
    \end{tabular}
    }
\end{table*}

\subsection{IoU-based Query Selection} 
In DETR, the tokens output by the decoder are matched one-to-one with ground truth targets via the Hungarian matching algorithm~\cite{kuhn1955hungarian}. During this process, the query selection schemes typically adopt a simple strategy, selecting the top-K queries with the highest scores. However, due to the inconsistency between classification scores and localization confidence, some predicted boxes with high classification scores are actually far from the ground truth boxes~\cite{zhao2024rtdetr}. As a result, boxes with high classification scores but low IoU are selected, leading to suboptimal detection results. To tackle this problem, we introduce IoU-based query selection, which encourages the model during training to assign high classification scores to features with high IoU scores and low classification scores to features with low IoU scores. We incorporate the IoU score into the objective function of the classification branch, similar to VFL~\cite{zhang2021vfl}, thus, the redefined optimization objective of the detector is:
\begin{equation}
\label{loss}
    \mathcal{L}(\hat{y}, y) = \mathcal{L}_{box}(\hat{b}, b) + \mathcal{L}_{cls}(\hat{c}, c ,IoU) ,
\end{equation}
where $\hat{y}$ and $y$ denote predicted result and ground truth, $c$ and $b$ denote categories and bounding boxes, and $\hat{y}=(\hat{c}, \hat{b})$, $y=(c,b)$.

\section{Experiments} 
\subsection{Dataset and Evaluation Metric}  
To validate the effectiveness and robustness of our method, we conducted comprehensive evaluations on several benchmark datasets:\\
$\bullet$ \textbf{{EvDET200K Dataset}} is a high-resolution event-based dataset for object detection, captured by PropheseeEVK4-HD sensor with a resolution of 1280$\times$720 pixels, containing 10 object categories in daily-life scenarios, with 10,054 event stream videos and approximately 200K bounding box annotations. \\
$\bullet$ \textbf{{DSEC Dataset}} is a stereo vision dataset designed for driving scenarios, captured under both favorable and challenging illumination conditions. The dataset includes 53 long sequences recorded during real-world driving in various lighting environments. \\ 
$\bullet$ \textbf{{GEN1 Dataset}} is captured using a Prophesee GEN1 event camera sensor with a resolution of 304$\times$240 pixels, mounted on a vehicle dashboard. It comprises 39 hours of real-world driving data, covering a wide range of scenarios including urban streets, highways, suburban areas, and countryside roads.

For evaluation, we employed the mean Average Precision (mAP) at various IoU thresholds, a standard metric widely used in object detection tasks. In addition, we evaluated the detector in terms of complexity and efficiency by measuring the number of parameters, FLOPs, and inference speed.

\subsection{Implementation Details} \label{detail}
Our code is implemented in Python using the PyTorch~\cite{paszke2019pytorch} framework. The experiments are conducted on a single NVIDIA A800 GPU. The model is optimized using the AdamW optimizer with an initial learning rate of 0.001 and weight decay of 0.0001, trained for 100 epochs. To achieve optimal training performance, we preprocess all input images by implementing comprehensive data augmentation strategies, such as photometric distortions, zoom out, IoU crop, and horizontal flip. The model is structured into four stages, with 2, 2, 12, and 2 CHCO layers stacked in each corresponding stage. These configurations enable efficient training while ensuring stable convergence. More details can be found in our source code.

\begin{table*}[!h]
\centering
\caption{Generalization results on GEN1 and DSEC datasets.} 
\label{tab_res_gener}
    \resizebox{\textwidth}{!}{ 
    \begin{tabular}{c|c|c|c|c|c}
    \hline 
        \#&Algorithm  &Publish  &Backbone  &\makecell{EvDET200K $\rightarrow$ GEN1 \\ ($mAP, mAP_{50}, mAP_{75}$)}  &\makecell{EvDET200K $\rightarrow$ DSEC \\ ($mAP, mAP_{50}, mAP_{75}$)} \\
        \hline 
        01 &DETR &ECCV 2020 &Transformer &16.9, 42.1, 21.6 &10.3, 26.6, 13.7  \\
        \hline
        02 &YOLOv10 &NeurIPS 2024 &CNN &17.2, 42.9, 22.0 &8.7, 23.5, 10.8 \\
        \hline
        03 &S5-ViT  &CVPR 2024  &Former+SSM  &20.7, 48.3, 24.9  &12.2, 31.7, 18.5 \\
        \hline 
        04 &SpikeYOLO  &ECCV 2024  &SNN  &16.5, 40.0, 20.1  & 9.0, 23.5, 10.9\\
        \hline 
        05 &vHeat  &CVPR 2025  &vHeat  &24.5, 50.3, 27.3  &11.6, 31.1, 17.9 \\
        \hline    
        06 &MvHeat-DET  &CVPR 2025  &MvHeat  &25.3, 52.3, 28.9  &11.2, 30.6, 16.8 \\
        \hline   
        07 &Ours  &-  &CvHeat  &25.5, 52.1, 29.2  &12.4, 31.6, 18.6 \\
        \hline 
    \end{tabular}
    } 
\end{table*}

\subsection{Comparison on Public Benchmarks} 
\noindent \textbf{Results on EvDET200K Dataset.} We compare our proposed method CvHeat-DET with a range of state-of-the-art detectors on the EvDET200K benchmark dataset, as shown in Tab.~\ref{tab:EvDET200Kresults}. CvHeat-DET achieves the highest performance across multiple metrics, with an mAP@50:95 of 53.6, mAP@50 of 80.9, and mAP@75 of 56.9, outperforming all existing methods by a clear margin. Notably, it surpasses recent transformer-based models such as Swin-T (49.0) and DETR (40.9), as well as event-specific approaches like SpikeYOLO (41.2) and EMS-YOLO (32.1). Despite the strong detection performance, the inference speed of CvHeat-DET is relatively slower compared to other methods. This is primarily due to the online graph construction mechanism adopted in our framework.

\noindent \textbf{Generalization Results on GEN1 and DSEC datasets.} As shown in Tab.~\ref{tab_res_gener}, we train several models on the EvDET200K dataset and evaluate their generalization performances on the Gen1 and DSEC datasets. As shown in the table, our method achieves better generalization compared to the baseline vHeat and other state-of-the-art detectors. These results further demonstrate the generalization capability of our proposed method, which can effectively adapt to detection tasks across diverse scenarios.

\subsection{Component Analysis}
The Tab.~\ref{tab_abl_component} presents a detailed component analysis on the EvDET200K dataset using various techniques and compares their performance against our baseline model. We use vHeat as our baseline, it achieves 50.0 mAP. When the IoU-based Query selection scheme (IQS) is introduced alongside the baseline model, the mAP score increases to 50.3. This indicates that IQS contributes positively to the model's performance. Furthermore, we utilize only the final Contour Graph for predicting the thermal diffusivity coefficient $k$, which results in an increase of the mAP score to 53.2. This indicates that the Contour Graph is capable of capturing more nuanced structural information within the images, aiding in the precise localization of objects. Additionally, by using graphs during the graph pruning process to form a multi-scale input, the model's mAP score is improved to 53.6, demonstrating the effectiveness of the multi-scale processing approach in graph structures.

\begin{table}[!h]
\centering
\caption{Component Analysis on EvDET200K dataset. We use vHeat as our baseline.}
\label{tab_abl_component}
\resizebox{0.48\textwidth}{!}{
    \begin{tabular}{c|cccc|c}
    \hline
        \textbf{Index} &\textbf{Baseline}  &\textbf{IQS}  &\textbf{Contour Graph}  &\textbf{Multi-scale}  & \textbf{mAP} \\
        \hline 
        1 &\checkmark  &  &  &  & 50.0 \\
        2 &\checkmark  &\checkmark  &  &  & 50.3 \\
        3 &\checkmark  &\checkmark  &\checkmark &  &53.2  \\
        4 &\checkmark  &\checkmark  &\checkmark  &\checkmark  &53.6  \\
    \hline
    \end{tabular}
}
\end{table}

\subsection{Ablation Study} 

\noindent $\bullet$ \textbf{Analysis on Different Input Formats.} 
Tab.~\ref{tab_abl_input} presents the results of different event data input formats, specifically Event Frames and Event Voxels. Using Event Frames yields a slightly higher mAP (53.6) compared to Event Voxels (52.7), indicating better detection performance. Moreover, Event Frames are significantly more efficient, requiring only 98.8G FLOPs and 1.27 seconds per iteration, whereas Event Voxels incur a much higher computational cost of 176.5G FLOPs and 5.36 seconds per iteration. These results suggest that, despite the finer temporal granularity offered by voxel representations, Event Frames strike a better balance between accuracy and computational efficiency. As such, we adopt Event Frames as the default input format in our framework.
\begin{table}[!h]
\centering
\caption{Results of different input formats.}
\label{tab_abl_input}
    \begin{tabular}{l|ccc}
    \hline
    Input Format & mAP & FLOPs & Speed \\ \hline
    Event Frames &53.6  &98.8G  &1.27 s / it  \\
    Event Voxels &52.7  &143.5G  &3.36 s / it  \\ \hline
    \end{tabular}
\end{table}

\begin{table*}[!h]
\centering
\caption{Results of different graphs. We use vHeat as our baseline.}
\label{tab_abl_graph}
\resizebox{0.85\textwidth}{!}{ 
\begin{tabular}{c|cccc|cc}
    \hline
        \textbf{Index} &\textbf{Baseline}  &\textbf{Global Graph}  &\textbf{Connected Subgraph}.  &\textbf{Contour Graph}  & \textbf{mAP} & \textbf{Speed}\\
        \hline 
        1 &\checkmark  &  &  &  &50.0  & 0.82 s / it\\
        2 &\checkmark  &\checkmark  &  &  &52.4 &2.07 s / it\\
        3 &\checkmark  &  &\checkmark &  &53.0  &1.26 s / it\\
        4 &\checkmark  &  &  &\checkmark  &52.8  &1.18 s / it\\
        5 &\checkmark  &\checkmark  &\checkmark  &\checkmark  &53.6 & 1.27 s / it\\
    \hline
\end{tabular}}
\end{table*}

\noindent $\bullet$ \textbf{Analysis on Thermal Diffusivity $k$.}
In physical heat conduction calculations, the thermal diffusivity $k$ plays a decisive role in the diffusion patterns. The Tab.~\ref{tab_abl_k} presents experimental results under different settings of $k$.
We conduct three sets of experiments to explore the influence of $k$:
First, $k$ is treated as a fixed hyperparameter. After testing various values, we select $k = 3$, which yield the best performance with an mAP of 49.7.
Second, $k$ is set as a learnable parameter and optimized during training. This approach achieves a higher mAP of 51.0.
Third, We further introduce the FEs to predict $k$, enabling the model to dynamically adjust the thermal diffusivity based on the input. This setting results in the best performance, with the mAP improving to 53.6.
These results demonstrate that more precise modeling of the thermal diffusivity $k$, especially through data-driven prediction, can significantly enhance the model's detection accuracy and generalization ability.

\begin{table}
\centering
\caption{Results of different thermal diffusivity \(k\). }
\label{tab_abl_k}
    \begin{tabular}{ll}
        \hline
        \textbf{Settings}            & \textbf{mAP} \\ \hline
        Fixed \(k\) = 3                  &49.7  \\
        \(k\) as learnable parameter &51.0  \\
        Predicting \(k\) using FEs   &53.6  \\ \hline
    \end{tabular}
\end{table}

\noindent $\bullet$ \textbf{Analysis on Different Graph.}
Tab.~\ref{tab_abl_graph} presents the ablation results on various graph designs. Starting from the baseline model (Index 1, mAP 50.0), we first incorporate a Global Graph (Index 2), which yields a notable improvement to 52.4 mAP. Introducing Connected Subgraphs alone (Index 3) further boosts performance to 53.0. These graph structures contain rich node information and effectively preserve global context, resulting in improved performance over the baseline. However, this comes at the cost of increased inference time. When combined with Contour Graphs (Index 4), the model achieves 52.8 and gets a faster speed, indicating that Contour Graph effectively reduces the graph scale while still preserving sufficient object boundary information. We integrate the three types of graph into the backbone in a multiscale manner, achieving a good balance between accuracy and efficiency.

\noindent $\bullet$ \textbf{Analysis on Different Resolutions of Event Stream.} 
In this section, we investigate the impact of event stream resolution on detection performance. Experiments are conducted with four different resolutions: 256$\times$256, 448$\times$448, 512$\times$512, and 640$\times$640 pixels. As shown in Tab.~\ref{tab_abl_res}, the detection results for 256$\times$256 resolution are 43.8/73.2/44.0, for 448$\times$448 are 49.9/78.6/52.4, for 512$\times$512 are 51.9/79.8/54.6, and for 640$\times$640 are 53.6/80.9/56.9. Higher-resolution event streams preserve more spatial information, which can positively influence model performance; however, they also lead to increased model complexity. Although the model does not perform particularly well at lower resolutions, it still achieves competitive results at intermediate resolutions, demonstrating its adaptability to scenarios with limited computational resources.

\begin{table}[!h]
    \centering
    \caption{Results of different input resolutions. }
    \begin{tabular}{l|cccc}
        \hline
        \textbf{Resolution}& \textbf{mAP} &\textbf{mAP@50} &\textbf{mAP@75} &\textbf{FLOPs}\\
        \hline
        256$\times$256px&43.8  &73.2  &44.0  &15.9G\\
        448$\times$448px&49.9  &78.6  &52.4  &48.7G \\
        512$\times$512px&51.9  &79.8  &54.6  &63.5G \\
        640$\times$640px&53.6  &80.9  &56.9  &67.7G \\
        \hline
    \end{tabular}
    \label{tab_abl_res}
\end{table}

\noindent $\bullet$ \textbf{Analysis on Number of CHCO in Each Stage.}
We experimented with stacking different numbers of CHCO modules to explore their impact on model performance. The Tab.~\ref{tab_abl_CHCO} presents the experimental results of using varying numbers of CHCO modules in different stages. The results indicate that increasing the number of stacked CHCO modules can significantly enhance the model's performance on detection tasks (53.1, 53.6, 53.7, 53.0). However, this also leads to a substantial increase in computational cost and parameter size. When we increased the configuration from (2, 2, 12, 2) to (2, 2, 18, 2), the performance improvement was only 0.1. Considering the limited gain in accuracy and the significant increase in computational resources, we adopted the (2, 2, 12, 2) configuration in our main experiments to achieve an optimal balance between accuracy and efficiency.

\begin{table}[!h]
    \centering
    \caption{Results of MHCO Number in Each Stage. }
    \label{tab_abl_CHCO}
    \resizebox{0.48\textwidth}{!}{
    \begin{tabular}{l|cccc}
        \hline
         \textbf{Number of MHCO}& (2,2,6,2) &(2,2,12,2) &(2,2,18,2) &(2,2,24,2)\\
         \hline
         \textbf{mAP}           &53.1       &53.6       &53.7       &53.0 \\
         \textbf{FLOPs}         &67.7G      &98.8G      &130.1G     &161.3G \\
         \textbf{Param}         &56.5M      &76.0M      &95.6M      &115.1M \\
        \hline
    \end{tabular}
    }
\end{table}

\begin{figure*}[!htp]
    \centering
    \includegraphics[width=\linewidth]{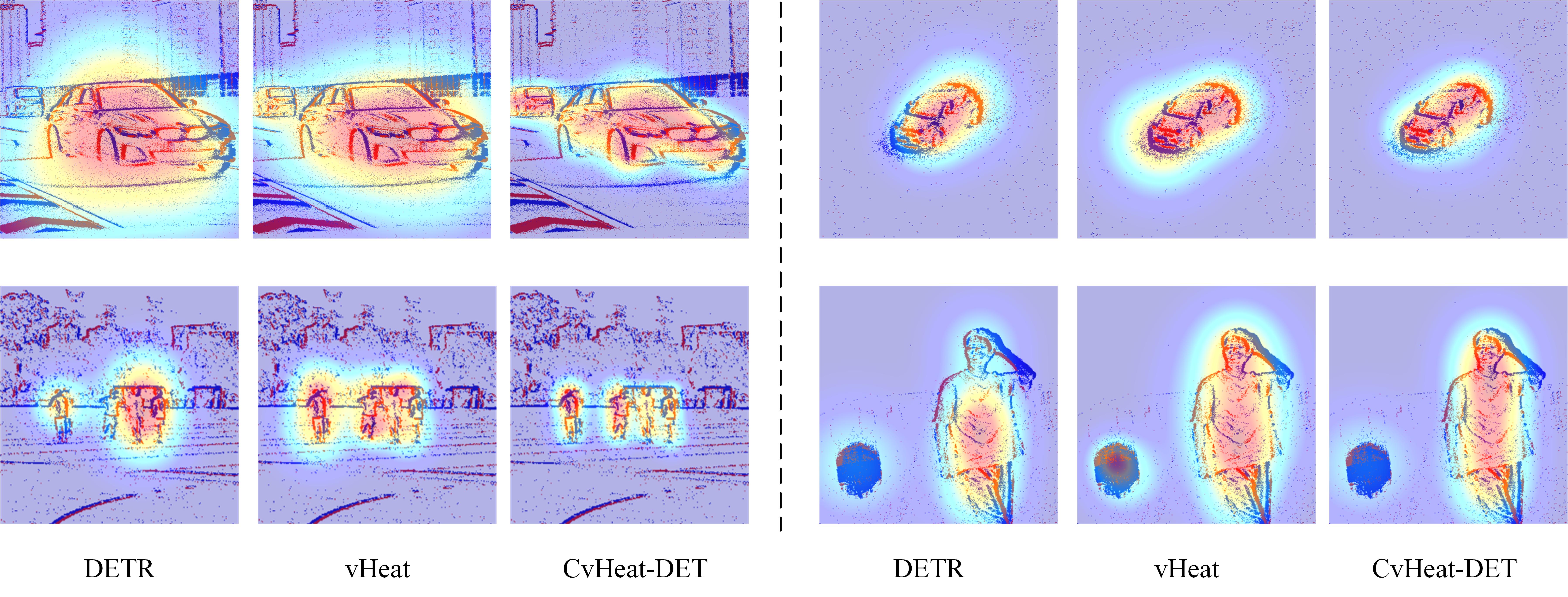}
    \caption{Visualization of the feature maps compared with other detectors.}
    \label{fig_featmap}
\end{figure*}

\begin{figure*}[!htp]
    \centering
    \includegraphics[width=\linewidth]{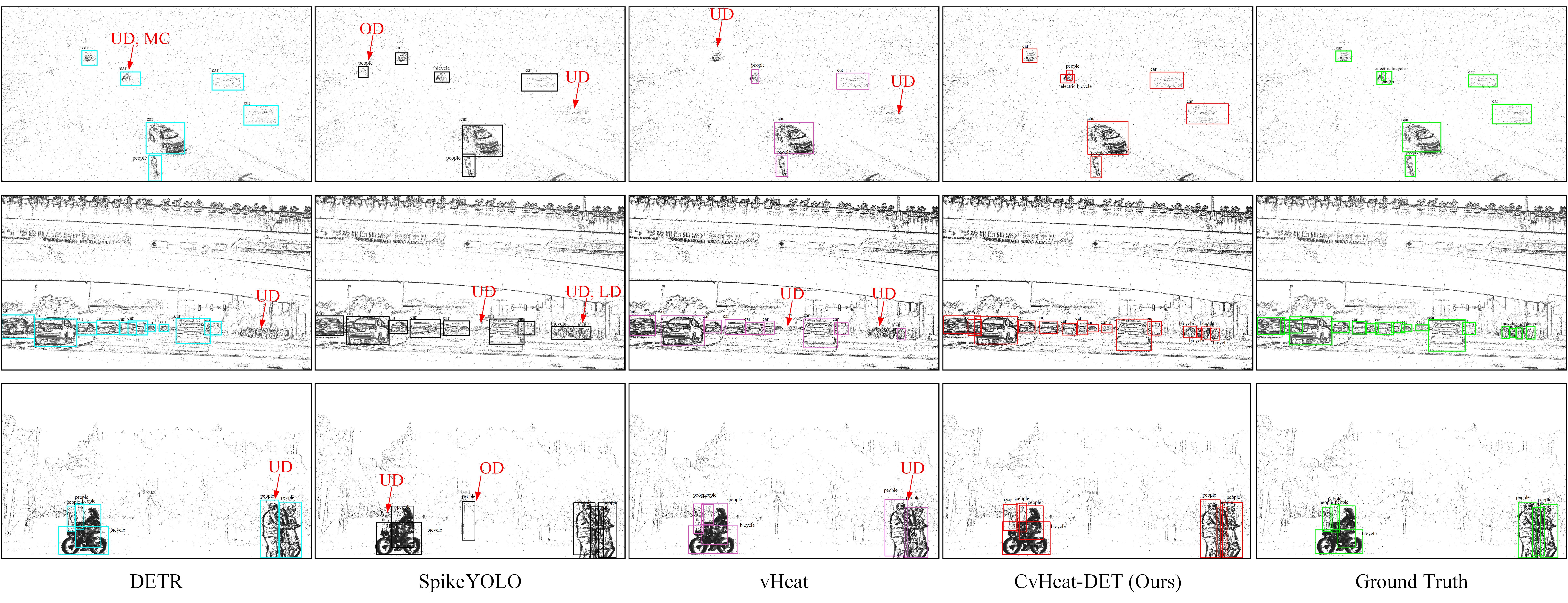}
    \caption{Visualization of the detection results of ours and other detectors. (MC: misclassification, UD: undetected, OD: over-detected, LD: large deviation.)}
    \label{fig_detres}
\end{figure*}

\subsection{Visualization} 
\noindent $\bullet$ \textbf{Feature Maps.}
Fig~\ref{fig_featmap} provides a comparative visualization of feature maps produced by three distinct detectors. Among these, our CvHeat-DET demonstrates a more precise focus on the objects of interest. Specifically, compared to the vHeat, CvHeat-DET exhibits a higher concentration of "heat", indicating more accurate and detailed object localization. This observation demonstrates the effectiveness of our proposed dynamic graph induced contour graph.

\noindent $\bullet$ \textbf{Detection Results.}
The comparison presented in Fig.~\ref{fig_detres} illustrates the detection results of our proposed CvHeat-DET model, alongside those of DERT, SpikeYOLO and vHeat. As shown in the figure, our detector exhibits robust performance even in dense scenes, while the other detectors struggle with missed detections or false positives in such challenging environments.

\subsection{Limitation Analysis}  
Although our proposed CvHeat-DET demonstrates strong object detection capabilities, the use of online graph construction and multi-scale graph features introduces limitations in training and inference speed. In addition, the current model does not fully exploit temporal information. In the future, we plan to design a temporal modeling module to integrate temporal cues, thereby enhancing the model's ability to perceive dynamic objects.

\section{Conclusion}  
In this paper, we propose CvHeat-DET, a novel dynamic graph-induced contour-aware heat conduction network tailored for event stream-based object detection. We use a heat conduction-based backbone to achieve a better balance between accuracy and efficiency. By introducing a dynamic structured graph, our model effectively captures the structural patterns in event streams and predicts thermal diffusivity coefficient $k$ to guide the heat conduction process. Furthermore, the incorporation of multi-scale feature representations enhances the network’s capability to model complex spatial structures. Experiments demonstrate that CvHeat-DET achieves superior performance compared to SOTA methods, highlighting its potential for robust and efficient event-based object detection in real-world scenarios.

{
    \small
    \bibliographystyle{ieeenat_fullname}
    \bibliography{main}
}


\end{document}